\documentclass[runningheads]{llncs}

\usepackage{graphicx}
\usepackage{amsmath}
\usepackage{tikz}
\usepackage{subfigure}
\usepackage{amsfonts}
\usepackage{multicol}
\RequirePackage{fix-cm}
\usepackage{threeparttable}
\usepackage{floatrow}
\usepackage[caption=false]{subfig}

\begin{document}

\title{Mapping the ocular surface from monocular videos with an application to dry eye disease grading}

\titlerunning{Mapping the ocular surface from monocular videos}

\author{Ikram Brahim\inst{1,2,3} \and
Mathieu Lamard \inst{1,3} \and 
Anas-Alexis Benyoussef \inst{1,3,4} \and  
Pierre-Henri Conze \inst{1,5} \and
B\'eatrice Cochener \inst{1,3,4} \and
Divi Cornec \inst{2,3} \and 
Gwenol\'e Quellec \inst{1}
}
\authorrunning{I. Brahim et al.}

\institute{LaTIM UMR 1101, Inserm, Brest, France \and
LBAI UMR 1227, Inserm, Brest, France \and 
University of Western Brittany, Brest, France \and 
Ophtalmology Department, CHRU Brest, Brest, France \and
IMT Atlantique, Brest, France
}

\maketitle              

\begin{abstract}

With a prevalence of 5 to 50\%, Dry Eye Disease (DED) is one of the leading reasons for ophthalmologist consultations. The diagnosis and quantification of DED usually rely on ocular surface analysis through slit-lamp examinations. However, evaluations are subjective and non-reproducible. To improve the diagnosis, we propose to 1) track the ocular surface in 3-D using video recordings acquired during examinations, and 2) grade the severity using registered frames. Our registration method uses unsupervised image-to-depth learning. These methods learn depth from lights and shadows and estimate pose based on depth maps. However, DED examinations undergo unresolved challenges including a moving light source, transparent ocular tissues, etc. To overcome these and estimate the ego-motion, we implement joint CNN architectures with multiple losses incorporating prior known information, namely the shape of the eye, through semantic segmentation as well as sphere fitting. The achieved tracking errors outperform the state-of-the-art, with a mean Euclidean distance as low as 0.48\% of the image width on our test set. This registration improves the DED severity classification by a 0.20 AUC difference. The proposed approach is the first to address DED diagnosis with supervision from monocular videos.

\end{abstract}

\section{Introduction}
 Dry Eye Disease (DED) is a condition that damages the ocular surface and tear film stability. DED can be traced back to a range of medical disorders, including Sjögren's syndrome, Parkinson, lupus, as well as smoking, contact lens, lasic surgery, or allergies \cite{manaviat2008prevalence}. One of the ways to assess the damaged ocular surface is through the staining \cite{wood2016diagnostic}. The tear quality and quantity can be  measured by the tear break-up time (TBUT), i.e. the interval between an eye blink and a tear break-up. Both ocular surface staining and TBUT have been used clinically for over a century \cite{begley2019review}. Algorithms have been proposed to automate the staining  (punctate dot grading) \cite{su2019superficial} and TBUT \cite{su2018tear} by analyzing digital slit lamp recordings and using supervised learning. But these tasks are still challenging due to eye motion and the video quality.
 \\
\indent One way to help facilitate DED quantification is through image registration, to compensate for eye motion and visualise the full eye. To achieve this goal we take into account multiple camera views and estimate the camera motion between them in an unsupervised manner using artificial intelligence (AI). The proposed method “SiGMoid: Semantic \& geometric monocular visual odometry”, learns how to predict both depth and egomotion, which allows for image registration. This method improves automated DED classification compared to a baseline.

\indent \textbf{Related works: Structure from motion (SfM)} is a complex problem in computer vision, whose aim is to reconstruct a 3D structure from a set of images. SfM uses images with different viewpoints in order to reconstruct a scene. More specifically, visual odometry (VO) is recovering the motion of a calibrated camera. This requires both camera motion estimation as well as inferred depth. The techniques differ in terms of what information is available as input \cite{aqel2016review}. Focusing on monocular visual odometry, the latest and most promising methods are Deep VO \cite{wang2020approaches}. For more precise estimations, this includes learning depth, optical flow, features and egomotion in a self-supervised manner. 
Unsupervised learning is made possible by a fundamental element, photometric loss, which is the difference between a pixel distorted from the source image, by estimated depth and pose, and the pixel recorded in the target view. Methods presented in \cite{casser2019struct2depth,casser2019unsupervised,zhou2017unsupervised} take advantage of structures and semantic segmentation for unsupervised monocular learning of depth and egomotion. Semantic image segmentation assigns a class to each pixel to indicate what is being represented. Semantics are used to identify moving objects and allow for robust egomotion estimation. The 3D geometry of the scene is used in \cite{mahjourian2018unsupervised} for more robust estimation. The method uses a 3D iterative closest point (ICP) loss, without prior shape knowledge, along with the photometric loss. In parallel, another team in \cite{wei2020semantics} explores the use of semantic segmentation of the scene for improvement as a novel 2D loss. They also combine a 3D ICP loss, which is less specific than a sphere fitting loss for our application. Lastly, a self-supervised spatial attention based depth and pose estimation is proposed by \cite{ozyoruk2020endoslam}. The method is applied for capsule endoscopy images and utilised synthetically generated data. 
\\
\indent These methods fall short for our objective because they are heavily driven by color, light disparity and shadows. In our examinations, just like many medical examinations, the source of light is attached to the camera. Light changes resulting from camera motion imply that matched points will have different colors in consecutive images. The resulting color variance is a disturbance and inhibits any learning through photometric loss, used in most methods. We propose a semantic segmentation reconstruction loss that disregards any change in color and ensures semantic constraint. This loss checks if matched points have the same semantic label in consecutive images. It should be noted that, in previous methods, semantic segmentation is only used to mask out moving/disturbing objects. Another limitation of previous methods is that they generally don't take advantage of the known geometry of the scene. In contrast, we propose a shape fitting loss, which penalizes unlikely depth maps, given the patient anatomy. This loss, which also relies on semantic segmentation, is made unique to our application as a sphere fitting loss. It is less complex than an ICP implementation yet influences more specific constraints.

\section{Proposed Method}
\begin{center}
    \begin{figure}
    \centering
\includegraphics[width=1\textwidth]{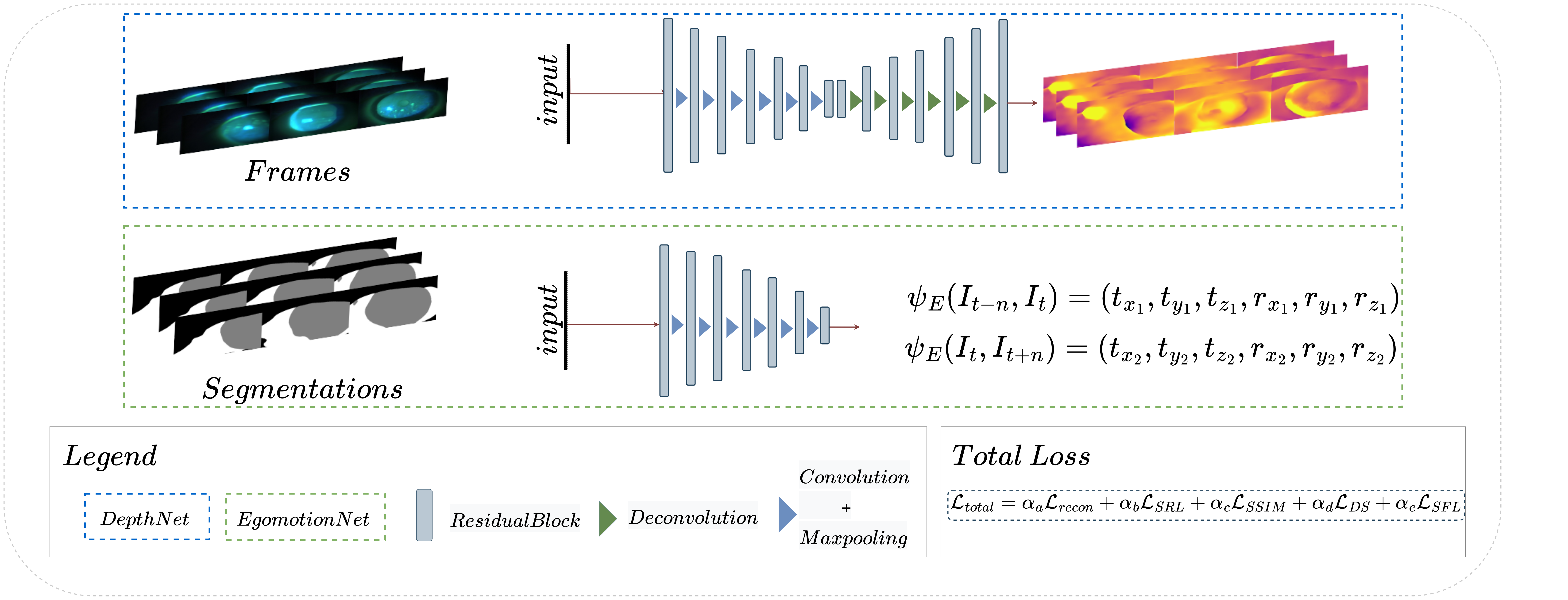}
\caption{SiGMoid framework.} \label{framework_sigmoid}
\end{figure}
\end{center}

We setup SiGMoid in the framework developed by Google, tested on autonomous driving datasets \cite{casser2019unsupervised,li2020unsupervised,gordon2019depth}.
Given the extracted frames from the examination videos, and the camera intrinsic parameters, we want to learn how to transform one frame onto another's coordinate system. It involves learning the depth of each frame and the egomotion between two. Our contributions are two new losses, semantic reconstruction loss and a sphere fitting loss. We first use a previously trained segmentation network (based on Feature Pyramid Networks - FPN \cite{lin2017feature}), to assign each pixel to the following classes: eyelid, sclera, cornea. All pixels labeled as ‘eyelid’ are ignored in all training and inference since they have no valuable information for the target applications (TBUT, punctate dot grading). The eyelid also moves with respect to the eyeball, and therefore violates the assumption behind the photometric and semantic losses, and cannot be modeled by a rigid (spherical) shape model. 
The predicted semantic segmentations are then used for both training and inference. Our framework, detailed below Fig.\ref{framework_sigmoid}, includes two CNNs (DepthNet and EgomotionNet) joined by the semantic reconstruction and photometric loss, which can be trained jointly (sharing of weights). 
Although trained jointly, they can be used separately for inference.
Depth is inferred by DepthNet using a single image and simultaneously the camera pose is computed from two frames in a sequence. 
Inputs to the CNNs are: Frames  \textit{I} : \([I_{t-n},I_t,I_{t+n}]\) and the Semantic Segmentation \textit{S} : \([S_{t-n},S_t,S_{t+n}]\) are used for loss calculations.

\begin{enumerate}
    \item Depth:
 a fully convolutional encoder-decoder architecture produces a depth map from a single RGB frame. 
 \begin{equation}
     \emph{D}_i=\theta(I_i)  , \theta  :  \mathbb{R}^{(H \times W \times 3)} \rightarrow  \mathbb{R}^{(H \times W)}
 \end{equation}
    \item Egomotion: a network takes three frames (ex. \([I_{t-n},I_t,I_{t+n}]\)) and predicts transformations simultaneously 
    \begin{equation}
        \psi_E(I_{i-n},I_{i}) = (t_{x_1},t_{y_1},t_{z_1},r_{x_1},r_{y_1},r_{z_1}) 
    \end{equation}

    \begin{equation}
         \psi_E(I_{i},I_{i+n}) = (t_{x_2},t_{y_2},t_{z_2},r_{x_2},r_{y_2},r_{z_2})
    \end{equation}
\end{enumerate}
The losses used to obtain the warped image, using a differentiable image warping operator $\phi(I_i, D_j, E_{i\rightarrow j})\rightarrow \hat{I}_{i\rightarrow j}$, make up the total loss for every pair of {$\hat{I}_{i \rightarrow j} - {I}_j$ }, Eq.\ref{total_loss} and are detailed below. The warping operator takes an RGB image  ${I}_{i}$ , a depth map  ${D}_{j}$, and the egomotion $E_{i\rightarrow j}$, giving us the reconstructed $\hat{I_{j}}^{th}$ image.

\begin{equation}\label{total_loss}
      \mathcal{L}_{total}= \alpha_a\mathcal{L}_{SRL} + \alpha_b\mathcal{L}_{recon} + \alpha_c\mathcal{L}_{SSIM} +
\alpha_d\mathcal{L}_{DS} +
\alpha_e\mathcal{L}_{SFL}
\end{equation}
\noindent\textbf{Semantic reconstruction loss \tiny(SRL)} is the main supervision signal.
\begin{equation}
    \mathcal{L}_{SRL}=min(\| \hat{S}_{i \rightarrow j} - {S}_j \|)
\end{equation}
\noindent\textbf{Photometric loss \tiny(RECON)} is similar to SRL except frames are used as input to compare this reconstructed image $\hat{I}_{i \rightarrow j }$ to the next frame ${I}_{j}$ \cite{casser2019struct2depth}.
\begin{equation}
   \mathcal{L}_{recon}=min (\| \hat{I}_{i \rightarrow j} - {I}_j \|)
\end{equation}
\noindent\textbf{Structural similarity loss \tiny(SSIM)} is used to assess the quality of the warping \cite{wang2004image}. SSIM combines comparisons of luminance, contrast and structure \cite{wang2004image}, and we measure this between $\hat{I}_{i \rightarrow j }$ and ${I}_{j}$. 
\begin{equation}
\mathcal{L}_{SSIM}= 1-SSIM(  \hat{I}_{i \rightarrow j} ,{I}_j  )
\end{equation}

\[SSIM( \hat{I}_{i \rightarrow j} ,{I}_j ) = \frac{(2\mu_{\hat{I}_{{i \rightarrow j}} } \mu_{{I}_j}  + l_1)(2 \sigma_{{\hat{I}_{{i \rightarrow j}} }{I}_j } + l_2)}{(\mu^2_{ \hat{I}_{i \rightarrow j} } + \mu^2_{{I}_j}  + l_1)
(\sigma^2_{\hat{I}_{i \rightarrow j}}+ \sigma^2_{{I}_j}  + l_2)} \\ 
\]
where $\mu_{ \hat{I}_{i \rightarrow j}},\mu_{{I}_j}$ are the average , $\sigma^2_{ \hat{I}_{i \rightarrow j}},\sigma^2_{{I}_j}$ the variance, and $\sigma_{{ \hat{I}_{i \rightarrow j}}{{I}_j}}$ covariance of  ${\hat{I}_{i \rightarrow j}},{I}_j$. $l_1 = (k_1 L)^2$,
$l_2 = (k_2 L)^2$, $L$ the dynamic range of the pixel-values, $k_1=0.01$, $k_2=0.03$.

\noindent\textbf{Depth smoothness \tiny(DS)} encourages smoothness by penalizing depth discontinuity if the image shows continuity in the same area \cite{godard2017unsupervised}. 
\begin{equation}
    \mathcal{L}_{DS}=|\nabla_x D_i |e^{-\nabla_x I_i}|  + |\nabla_y D_i |e^{-\nabla_y I_i}|
\end{equation}
where $\nabla_x,\nabla_y$ are image gradients in the horizontal and vertical direction, respectively.

\noindent\textbf{Sphere fitting loss \tiny(SFL)} is implemented for what is estimated to be the eye's shape: two intersecting spheres \cite{park2018learning}. A smaller anterior transparent sphere is the cornea and a posterior sphere representing the sclera. In order to implement this loss, we first estimate a depth map, and then using the semantic segmentation, we calculate the sphericity of the two regions: cornea and sclera.
As detailed in \eqref{sphere_fitting} we first use the segmentations to obtain the depth predictions of either regions. The sphere fitting is then implemented twice for both, the loss is the sum of both errors. We apply a threshold of 0.5 before calculating this loss to ensure either regions are present in the frame. We define our threshold as the count of non zero pixels pertaining to either regions and dividing that by the total number of pixels of the frame. 
Once a frame's calculated region presence exceeds the threshold,  we use the depth estimations for either the corneal or scleral region and convert it to a 3D point cloud projection using the inverse of the intrinsic matrix. We then use the estimated point cloud and apply a least squares sphere fitting. Following the method proposed by Jekel, we are able to determine the best sphere center for the given data points \cite{Jekel2016}. By rearranging the terms in Eq. \ref{sphere_eq}, we can express the equation in matrix notation and solve for $\vec{c}$ (see Eq.\ref{sphere_solve}). By fitting the $n$ data points $x_k,y_k,z_k$, we can solve for the centre coordinates of the sphere $x_0,y_0,z_0$ and the radius $r$. 

\begin{equation}\label{sphere_eq}
    (x - x_0)^2 + (y - y_0)^2 + (z - z_0)^2 = r^2
\end{equation}
\begin{equation}\label{sphere_rearranged}
    x^2 + y^2 + z^2 = 2xx_0 + 2yy_0 + 2zz_0 + r^2 - x_0^2 - y_0^2 - z_0^2
\end{equation}
\begin{equation}\label{sphere_solve}
    \vec{f} = A\vec{c}
\end{equation}

\begin{equation}
  \vec{f}= \begin{bmatrix}
  x_k^2 + y_k^2 + z_k^2 \\
  x_{k+1}^2 + y_{k+1}^2 + z_{k+1}^2 \\
  \vdots \\
  x_{n}^2 + y_{n}^2 + z_{n}^2
 \end{bmatrix} 
 A = \begin{bmatrix}
  2x_k & 2y_k & 2z_k & 1 \\
  2x_{k+1} & 2y_{k+1} & 2z_{k+1} & 1 \\
  \vdots & \vdots & \vdots & \vdots \\
  2x_n & 2y_n & 2z_n & 1 \\
 \end{bmatrix} 
 \vec{c} = \begin{bmatrix}
  x_0 \\
  y_0 \\
  z_0 \\
  r^2 - x_{0}^2 - y_{0}^2 - z_{0}^2
 \end{bmatrix} 
\end{equation}

Sphere fitting loss $\mathcal{L}_{SFL}$ is a mean square error (MSE) between the fitted sphere and the data points. The sphericity for each of the corneal and scleral regions have a weight $\alpha_e$. With both regions fitted to a sphere we then calculate the loss for each pixel $p$.
\begin{equation}\label{sphere_fitting}
  \mathcal{L}_{SFL}= \mathcal{L}_{cornea_{SFL}} + \mathcal{L}_{sclera_{SFL}}
\end{equation}
\begin{equation}\label{sphere_cornea}
\mathcal{L}_{cornea_{SFL}}= \frac{1}{p_c} \sum_{k=1}^{p_c} ((x_{ck}-x_{c0})-r_c)^2 ,   
\mathcal{L}_{sclera_{SFL}}= \frac{1}{p_s} \sum_{k=1}^{p_s}  ((x_{sk}-x_{s0})-r_s)^2
\end{equation}

where $p_c$ are pixels, $x_{ck}$ data points, $x_{c0}$ centre coordinate on the corneal surface, $r_c$ the cornea radius and $p_s$ are pixels, $x_{sk}$ data points, $x_{s0}$ centre coordinate on the scleral surface, $r_s$ the estimated sclera radius.

\section{Experiments \& Results}
\textbf{Dataset}
The dataset was collected from slit lamp videos taken during the examination of patients with Sjögren's syndrome (PEPPS study). This is a prospective cohort evaluating the ocular surface damages in patients with Sjögren's syndrome followed at the university hospital of Brest. The videos were recorded using the Haag Streit BQ 900 slit lamp and the camera module CM 900 (resolution 1600$\times$1200, 12 fps, magnification $\times$10). The ocular surface was analysed after illumination with white light (lissamine green evaluation) followed by cobalt blue light and interposition of a yellow filter (fluorescein evaluation). Our database contains 26 videos from 26 patients.

\subsection{SiGMoid}
Using our dataset, we conducted several experiments mainly using transfer learning. We used models pre-trained on the Cityscapes dataset (C) \cite{cordts2016cityscapes,gordon2019depth}.  

\noindent\textbf{Preprocessing}
Calibration was performed using Matlab (MathWorks, Natick, MA) with images of a planar checkerboard (8 $\times$ 7 squares of 2 $\times$ 2 mm). We employed 10 calibration images (pattern placed at different poses). The intrinsic parameters obtained from the calibration were the focal length fx=fy=3758.9 pixels, and the central points cx = 138.8, cy=85.4.

To prepare the data we first use our trained FPN model to predict the semantic segmentation for each of the frames.
For training we produce three-frame sequences with an interval of $n=10$ frames. This is a setup we chose given that our videos have 12 frames per second and the motion between consecutive frames is usually small. This resulted in 15,275 three-frame sequences for training.
\\
\textbf{Evaluation}
Unlike existing methods, we do not have access to ground truth depth and odometry.
We manually annotated punctate dots (damaged areas) on the surface of the eye, and visible veins on the sclera. To visualise the accuracy of our predictions, we warp a source frame into a target frame and tracked the marked points. By using the equation $p_{s} \sim K \hat{T}_{t\xrightarrow{}s} \hat{D}_t(p_t) K^{-1}p_{t}$, we project source frame pixels $p_s$ to the target frame $p_{t}$. Our evaluation used the inverse warping which first requires the depth map $\hat{D}_t$ prediction of the target frame, and then the egomotion from source to target which also gives us the transformation matrix $\hat T_{t \rightarrow s}$. 
Evaluation was performed on a test set of 3 patients with 54 frames. Our test set consists of 126 points on the sclera on vein intersections and punctate dots, 33 points on the cornea of visible punctate dots. 
\\
\textbf{Training setup} The baseline we used to compare is the implementation of \cite{casser2019unsupervised}, which had to be trained in intervals due to the loss diverging to infinity, as well as the predicted egomotion matrix being non-invertible.
Baseline and SiGMoid were trained using a learning rate 0.0002, batch size of 8, SSIM weight 0.15, and depth smoothing weight 0.04 \cite{casser2019struct2depth}.
The baseline L1 reconstruction weight 0.85 and SiGMoid's training setup was using L1 reconstruction weight 0.15, L2 semantic reconstruction loss weight 0.85, sphere fitting weight 10000 (given the very small distances this required a big weight). All models were trained until 200 epochs, with the best model (lowest loss achieved) saved for inference.
\\

\noindent\textbf{Results} Table \ref{table_2} shows the evaluation results on our test set by different methods. By defining the two novel losses, we are able to stabilise training and avoid the loss diverging and obtain better results. Despite the increase in computation during training via the losses, our inference time remains similar to the baseline. In the following experiments, we preprocessed the data using a frame step $n=10$. 
We compared two configurations: SiGMoid no.$3$ with only $\mathcal{L}_{SRL}$, SiGMoid no.$4$ with both contributions $\mathcal{L}_{SRL}$ and $\mathcal{L}_{SFL}$.

\begin{figure}[!h]
\centering
   \begin{floatrow}
     \subfigure[Frame]{\includegraphics[width=0.2\textwidth]{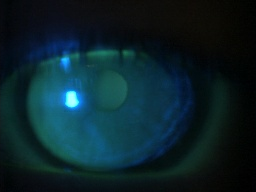}}\hfill
     \subfigure[ \scriptsize{Segmentation}]{\includegraphics[width=0.2\textwidth]{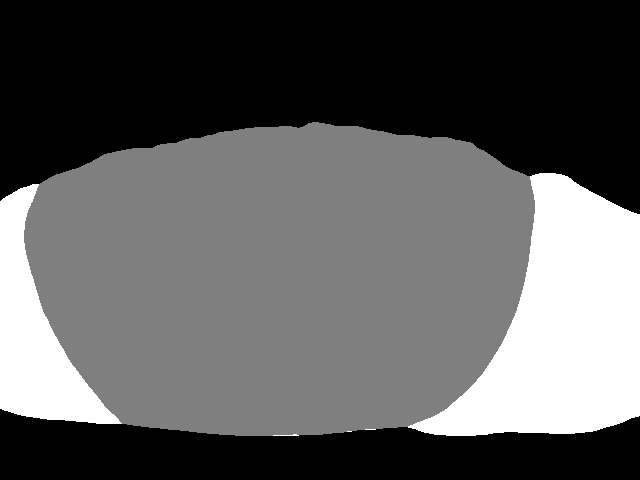}} \hfill
    \subfigure[Casser \textit{et al.} \cite{casser2019unsupervised}]{\includegraphics[width=0.2\textwidth]{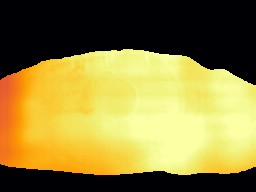}} \hfill
    \subfigure[SiGMoid 4]{\includegraphics[width=0.2\textwidth]{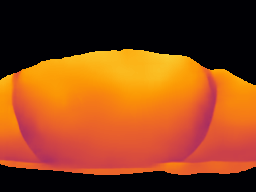}}\hfill
   \end{floatrow}
   \caption{CNN inputs \& Depth map predictions.}
   \label{fig:depth_preds}
\end{figure}

The mean Euclidean distance in pixels is lowest in SiGMoid no.$4$ with both CNN inputs being the frames. Our proposed method achieves the lowest error when compared to the baseline. The reconstruction improves in all SiGMoid implementations proving that a simple photometric loss used in \cite{casser2019unsupervised} is less efficient for our data. We also compute a mean inter-grader error (result no. 5). We asked three graders to re-annotate the test set points giving us an average human error of 4.81 px. We visualise the depth map predictions in Fig.\ref{fig:depth_preds}.

\begin{center}
\begin{table}[!h]
\caption{Experiment details and results,.}\label{table_2}
\begin{center}
\begin{tabular}{c|c|c|c|c|c|c}
\hline
No. &

Method                                                                       & \begin{tabular}[c]{@{}c@{}}Frame\\ step\end{tabular} & $\mathcal{L}_{SRL}$ & $\mathcal{L}_{SFL}$  & \begin{tabular}[c]{@{}c@{}}Mean \\ Euclidian (px)\end{tabular} & \begin{tabular}[c]{@{}c@{}}Mean\\  Euclidan (\%)\end{tabular} \\ 
\hline

1 & Casser \textit{et al.} \cite{casser2019unsupervised}*                                                                          & No                                                    & -                              & -                             & 29.08                                                              & 1.82                                                             \\ \hline

2 & Casser  \textit{et al.} \cite{casser2019unsupervised}*                                                                          & Yes                                                   & -                              & -                             & 27.19                                                          & 1.70                                                          \\ \hline

3 & SiGMoid no.  3                                                                                                                                            & Yes                                                   & Yes                            & No                            & 22.48                                                          & 1.40                                                          \\ \hline
4  &SiGMoid   no. 4                                                                                                                                        & Yes                                                   & Yes                            & Yes                           & \textbf{7.7}                                                            & \textbf{0.48}                                                          \\ \hline

5  & Grader errors                                                                                                                                         & NA                                                   & NA                            & NA                           & 4.81                                                          & 0.30                                                        \\ \hline

\end{tabular}
\end{center} 
\begin{tablenotes}
    \item CNN Inputs \(\rightarrow\) DepthNet : Frames (I) ,  EgomotionNet : Frames (I)
    \item[*] denotes trained in multiple intervals due to loss divergence \( \rightarrow \infty\)

\end{tablenotes}
\end{table}

\end{center}

\subsection{DED Diagnosis: Classification}
In order to evaluate our proposed method in a diagnostic aspect, we apply it to the automated classification of mild versus severe DED. Mild DED (respectively severe DED) is defined as a corneal Oxford score  $\leq 1$ (respectively $\geq 2$) \cite{bron1997reflections}.

\noindent\textbf{Preprocessing}
We train the baseline with the raw frames from the videos of the examinations. We compare this with a mosaic from a pair of frames obtained using our registration method and selecting only the frames with a warping error (SRL) less than 5\%. This resulted in $\approx$ 35\% of frames being removed.

\noindent\textbf{Training Setup}
The training setup was identical for both experiments; learning rate 2e-06, batch size of 64 and using resnet50 as backbone. Due to data scarcity, an additional dataset of 28 videos acquired using a different examination protocol was used as the validation set for our DED classification training.
The data was split into ; 35 eyes for the train, 39 for the validation and 14 for the test. All eyes from the same patient were assigned to the same set.

\noindent\textbf{Evaluation}
Area under the (receiver operating characteristic) curve (AUC), accuracy (ACC), precision and recall were used as metrics. The validation and test evaluation were performed per eye by using a majority vote for the \textit{n} frames per patient. As shown in Table \ref{table_3}, the classification improves when using the registered frames obtained using SiGMoid. All metrics improve with a margin of 0.08-0.22 validating the application of our proposed method to DED grading.

\begin{table}[!h]
\begin{center}
\begin{tabular}{c|c|c|c|c|c|c}
\hline
No. & Method & Backbone & AUC & ACC & Precision & Recall \\
\hline
1 & Baseline  & resnet50 \cite{he2016deep} & 0.69 & 0.57 & 0.73    & 0.67      \\ \hline

2 & SiGMoid    & resnet50 \cite{he2016deep} & \textbf{0.89} & \textbf{0.79} & \textbf{0.81}   & \textbf{0.83} \\ \hline

\end{tabular}
\end{center}
\caption{Classification evaluation results}\label{table_3}

\end{table}

\section{Discussion and Conclusion}

We proposed SiGMoid, a self-supervised image registration algorithm towards DED diagnosis and quantification from slit lamp videos. This is the first use, to our knowledge, of monocular DED examinations in a self-supervised manner for this application. Our results validate that, due to the color/illumination variance present in the examinations, the baseline method is not sufficient. Although both contributions improved our results, we see a more significant improvement from the sphere fitting loss. Our method also has the closest mean euclidean distance to what we considered human error. Additional data acquired from a different acquisition device could enable to robustify our approach. In particular, it could allow us to test how generalizable our method is and also expand it through fine-tuning. Finally, we demonstrated that obtaining an accurate reconstruction is beneficial to the classification of DED grading. 

\section*{Acknowledgments}
This research was supported by funding from the Innovative Medicines Initiative 2 Joint Undertaking (JU) under grant agreement No 806975. The JU receives support from the European Union’s Horizon 2020 research and innovation program and EFPIA. It is also funded in part by The Brittany Region through the ARED doctoral program.

\bibliographystyle{splncs04}
\bibliography{mybibliography}
\end{document}